# Occlusion-aware Driver Monitoring System using the Driver Monitoring Dataset

Paola Natalia Cañas
*Connected & Cooperative
Automated Systems
Fundacion Vicomtech, (BRTA), and
University of the Basque Country*
San Sebastián, Spain
pncanas@vicomtech.org

Alexander Diez
*Faculty of Computer Science
University of the Basque Country*
San Sebastián, Spain
adiez134@ikasle.ehu.eus

David Galvañ
*Connected & Cooperative
Automated Systems
Fundacion Vicomtech, (BRTA)*
San Sebastián, Spain
dagalvan@vicomtech.org

Marcos Nieto
*Connected & Cooperative
Automated Systems
Fundacion Vicomtech, (BRTA)*
San Sebastián, Spain
mnieto@vicomtech.org

Igor Rodríguez
*Faculty of Computer Science
University of the Basque Country*
San Sebastián, Spain
igor.rodriguez@ehu.eus

*Abstract*—This paper presents a robust, occlusion-aware driver monitoring system (DMS) utilizing the Driver Monitoring Dataset (DMD). The system performs driver identification, gaze estimation by regions, and face occlusion detection under varying lighting conditions, including challenging low-light scenarios. Aligned with EuroNCAP recommendations, the inclusion of occlusion detection enhances situational awareness and system trustworthiness by indicating when the system's performance may be degraded. The system employs separate algorithms trained on RGB and infrared (IR) images to ensure reliable functioning. We detail the development and integration of these algorithms into a cohesive pipeline, addressing the challenges of working with different sensors and real-car implementation. Evaluation on the DMD and in real-world scenarios demonstrates the effectiveness of the proposed system, highlighting the superior performance of RGB-based models and the pioneering contribution of robust occlusion detection in DMS.

## I. Introduction

In recent years, significant advancements in navigation and autonomous control systems have enabled vehicles to perform various tasks and operate in specific environments autonomously. These systems, classified as Level 3 and above according to the SAE [1], allow drivers to delegate certain responsibilities to the automated system, effectively becoming passengers under certain conditions. However, even in autonomous mode, the system may require the driver's intervention in situations of uncertainty or exceptional circumstances. This transition of roles and control must be managed safely to ensure the driver's readiness to take over when necessary. Additionally, in lower levels of autonomy (e.g. < SAE L3), monitoring the driver is essential during manual driving to prevent human errors.

This is where driver monitoring systems (DMS) become vital. These systems assess the driver's attention, alertness, and readiness to take control, thereby enhancing overall safety.

Driver monitoring systems focus on collecting relevant information about the vehicle's cabin environment and the driver's physical and mental state. By measuring variables such as head pose, body pose, blinking behaviours, and other physiological signals, these systems can identify states of fatigue, distraction, and the driver's comprehension of vehicle alerts, among others. According to EuroNCAP, a driver's attention is primarily determined by the direction of their gaze, with distraction being defined as the driver not looking at the road. EuroNCAP has also established other guidelines for DMS (i.e. driver fatigue detection and how to estimate it) and appropriate vehicle responses and alerts [2] to ensure their effectiveness and reliability.

In addition to monitoring the driver, modern DMS now include functionalities such as driver identification. This feature enhances both security and comfort within the vehicle. For security purposes, driver identification can prevent unauthorized access and theft. For instance, facial recognition technology can ensure that only authorized individuals can start the vehicle. On the comfort side, driver identification allows for the automatic adjustment of seat positions, mirror angles, climate control, and infotainment preferences based on the identified driver. These personalized settings improve the driving experience by tailoring the vehicle environment to individual preferences. As a result, such applications are gaining relevance in the automotive industry, with additional innovative uses continually being developed.

DMS that rely on visual sensors, such as cameras, can experience performance degradation or prediction failures due to occlusions. Occlusions, in this context, refer to visual obstructions of the driver's face and eyes, which are the primary objects of analysis. These obstructions can be caused by various factors, including environmental elements like direct sunlight,

This work was funded by the Horizon Europe programme of the European Union, under grant agreement 101076868 (project AWARE2ALL). Funded by the European Union. Views and opinions expressed here are however those of the author(s) only and do not necessarily reflect those of the European Union or CINEA. Neither the European Union nor the granting authority can be held responsible for them.

XXX-X-XXXX-XXXX-X/XX/$XX.00 ©20XX IEEE

glare or shadows, personal accessories such as sunglasses or hats, facial hair like beards or moustaches, and heavy makeup. Also, depending on where the camera sensor is placed there is a risk of occlusion by the hand of the driver. Such occlusions can partially or completely obscure the driver's facial features, affecting the DMS algorithms' ability to accurately detect unsafe driving situations. To address this challenge, EuroNCAP recommends that these systems should either be robust against occlusions or alert the user when they are not operating correctly [2]. In addition, DMS must work at day and night lighting conditions. This is a special challenge for computer vision-based algorithms.

In this research, our objective is to develop a DMS that incorporates a reliable gaze estimation and driver identification algorithm. This system aims to be robust enough to function effectively under varying lighting conditions and to detect occlusions. By identifying when an occlusion occurs, the system can alert the driver to potential malfunctions, ensuring continuous and accurate monitoring. This document details the following developments:

- The creation of a gaze estimation by regions algorithm, an occlusion classifier and a driver identification
- All of the previous algorithms are prepared to use RGB and IR images. Relevant for driving lighting variation.
- The integration of these three algorithms into a cohesive system logic to ensure continuous and reliable functioning in real-world conditions.

*A. Gaze estimation based on regions*

Gaze detection systems operate by identifying the driver's gaze direction to estimate where they are looking at any given moment. This information provides valuable insights into the driver's cognitive state and their engagement with the driving task. Knowing this can provide information about the driver's distraction level [3], awareness of other road actors [4], and predicting driver's maneuvers [5].

Various methods have been employed for gaze estimation. Some techniques involve estimating eye vectors to precisely pinpoint the driver's focus resulting in a point (x,y,z) in the 3D space [6], often requiring rigorous calibration, while others identify the region of interest where the driver's attention is directed [7]. These methods often rely on analyzing the driver's head pose and eye appearance, using image processing and computer vision algorithms to extract relevant features and infer gaze direction [8].

In this research, although there are more sophisticated and accurate methods to perform gaze estimation, we developed a driver's gaze estimation algorithm based on gaze regions, utilizing both RGB and IR images to enhance robustness under varying lighting conditions and is integrated into an occlusion-aware system to prevent incorrect gaze outputs.

*B. Driver identification*

Driver identification is implemented to personalize car features [9] such as setting preferred seat positions, adjusting rear-view mirrors, climate settings, and even suggesting routes based on past destinations [10] once the driver is re-identified. Furthermore, the driver ID can unlock new third-party applications, including payments [11] or assurance services [12]. In the context of fleet management, driver identification proves useful for tracking work schedules, among other applications [13].

The identification of drivers has been achieved through the analysis of various signals, including image [14], voice [15], or driving behaviours [16]. However, our approach focuses on analyzing the driver's face for identification. The system developed in this research serves as a foundation for other developments, primarily related to comfort. While it is not crucial to implement additional measures to prevent spoofing [17] or other identity theft violations in this context, we prioritize the precision of re-identifying registered drivers. Additionally, we ensure the image of the driver is not degraded to be correctly compared against registered IDs, avoiding false rejections.

*C. Occlusion detection*

Infrared-only driver monitoring systems offer an advantage in handling varying lighting conditions due to their visibility in darkness. However, the absence of color in IR images leads to a loss of relevant information. Given that most available datasets consist primarily of RGB images, prioritizing the development and testing of RGB-based systems is crucial for creating more representative solutions. Nevertheless, IR-based systems remain essential for low-light scenarios and will serve as a fallback when RGB-based systems are ineffective.

Our research aims to create a resilient system capable of managing difficult lighting conditions, including dark environments where conventional techniques face challenges due to reduced visibility and poor image quality. This will involve adapting existing models to process infrared images effectively. Furthermore, the system will detect occlusions that could impair the performance of other algorithms, alerting the driver when it cannot function correctly, as recommended by EuroNCAP. To the best of our knowledge, there is no academic research related to occlusion detection from images for driver monitoring systems. This is the first attempt to introduce this functionality in a DMS to ensure proper functioning.

II. DATA

The data used for this research is entirely from the Driver Monitoring Dataset (DMD) [18]. With previous explorations of the data of distraction [19], this document presents the first analysis of gaze-related data and occlusion labels within the DMD. The DMD is a public video dataset containing recordings of 37 drivers engaged in various activities while driving. Annotations cover several aspects, including distraction, drowsiness, hand and wheel interaction, and gaze estimation. The recordings were captured by three cameras focusing on the driver's face, body, and hands, with each camera recording in RGB, IR, and depth data.

*A. Gaze data pre-processing*

The gaze-related data in the dataset includes annotations for blinks and gaze regions. For this research, we focus solely on the gaze region annotations. The DMD defines nine possible gaze regions, as illustrated in Fig. 1.

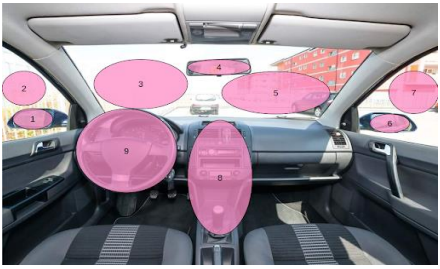

Fig. 1. Gaze regions of the DMD

Since the DMD dataset is in video format, we first extracted every frame along with its annotation. This process resulted in 221,172 RGB images and their corresponding IR counterparts. These sets of images contained significant redundancy, as contiguous frames were often nearly identical. To mitigate potential overfitting issues related to redundancy, we deduplicated the datasets. We achieved this by calculating the dHash perceptual hash for each image and removing those with identical hash values.

Upon inspecting the deduplicated images, we observed that some annotations did not correspond to the actual gaze region the person was looking at. This discrepancy primarily occurred in frames where individuals were transitioning between gaze zones. We manually removed all problematic images, resulting in a final total of 6,424 images in each dataset. Fig. 2 (a) displays some images from the pre-processed RGB and IR dataset.

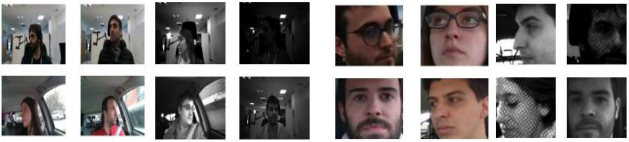

(a)                                              (b)

Fig. 2 RGB and IR image samples of DMD (a), and RGB and IR cropped image samples (b)

These images were divided into three groups: test, train, and validation, which accounted for 20.73%, 63.39%, and 15.88% of the dataset, respectively. We ensured that images of the same person did not appear in more than one group to prevent overconfidence in our models.

### B. Occlusion data pre-processing

The images for the gaze classifiers were sourced from the gaze-annotated part of the DMD, while the images for the occlusion classifiers were taken from the distraction-annotated part of the DMD, as they contained a higher number of occluded images. The vast majority of images in both the IR and RGB datasets correspond to the negative class, or the "no occlusion" class, resulting in highly unbalanced datasets. The images were divided into test, train, and validation sets, accounting for 20%, 64%, and 16% of the dataset, respectively. Class ratios were maintained across all splits to ensure consistency.

### III. GAZE CLASSIFIER

The initial RGB model experiments involved training a mobilenet_v3_small [20] architecture with PyTorch's IMAGENET1K_V1 pre-trained weights and with the classification layers modified to have 9 output neurons, corresponding to the 9 gaze zones. This architecture was chosen for its small size and low GFLOPS, enabling real-time inference. However, the model underfitted the data, achieving only 59.12% on test data. Consequently, a slightly larger model, Efficientnet_b0 [21] with PyTorch's pretrained weights, was employed. A baseline accuracy of 74.31% on the test set was achieved, indicating the model's suitability for the task. Further improvements were achieved by resizing input images to (256,256) and then performing a center crop of size (224,224), inspired by PyTorch's inference transforms for EfficientNet_b0. This increased accuracy to 79.12%, possibly due to the increased size of faces in the cropped images, improving the signal-to-noise ratio. Ultimately, cropping images with a face detector and passing only that to the classifier resulted in a test accuracy of 84.31%. MTCNN [22] was used for face detection due to its high accuracy and fast inference times. Fig. 2 (b) shows some samples of the cropped dataset.

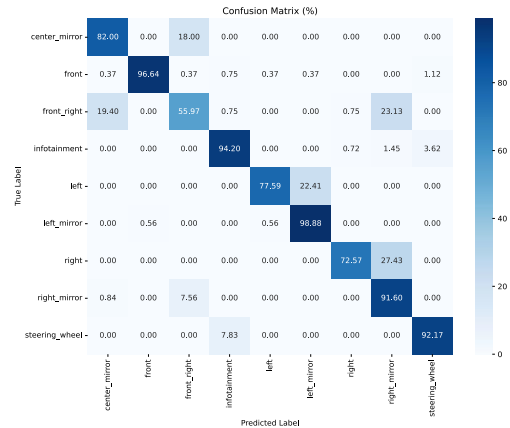

Fig. 3 Confusion matrix of RGB region gaze estimation algorithm

Final tuning of the learning rate and weight decay, along with the use of AugMix [23] and TrivialAugment [24] for data augmentation, resulted in a final test accuracy of **86.29%**. It can be seen that the model has trouble differentiating between the side mirrors and the windows on both sides, as well as confusing the front right with center and right mirror regions. This was expected due to the small degree of turn the driver must make to look at one of these regions or the other.

The IR model was initially trained using the same methods as the RGB model, which had produced good results. This involved cropping the IR dataset to only include faces. The resulting accuracy was 84.70%. However, real-world testing showed poor performance. Closer examination of the dataset revealed images with extreme lighting conditions, some too dark to see the face and others too bright to distinguish anything, as represented in Fig. 4. To address this, images with a mean pixel brightness above 235 or below 20 were removed, eliminating most of the extreme cases. The model was retrained on the new dataset, which still contained images with varied lighting, but less extreme. The test accuracy was **76.76%**. Despite the lower accuracy compared to the previous model, real-world performance was significantly better due to the training data being more representative of real-world conditions. This model was ultimately chosen for the final pipeline.

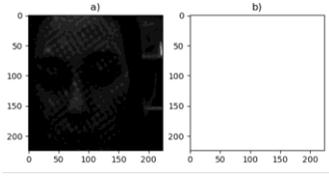

Fig. 4 IR images with extreme lighting issues examples

The difference in accuracy between the RGB and IR models is likely due to lighting issues affecting the IR model's learning and the quality of the IR data, which is highly dependent on the sensor used. A robust sensor that captures clear images in low-light conditions is crucial for good model performance.

## IV. OCCLUSION CLASSIFIERS

For RGB images, we initially used the same training strategy as the gaze classifier, as we believed it could achieve good accuracy. We trained a Mobilenet_v3_small initialized with PyTorch's IMAGENET1K_V1 weights and changed the classification layer to 2 output neurons. We then made gradual adjustments to the training to enhance performance. The final process used TrivialAugment as well. Still, it differed from the gaze classifier's regarding weight decay, learning rate, exclusion of AugMix, and a three-layer classification layer (instead of one). This model achieved an accuracy of 99.67% and, more importantly, an occlusion recall of 95.34%. Recall is crucial in this scenario due to the limited number of occlusion images; it indicates the percentage of those images that were correctly predicted.

While this model performed well on the test data, we cropped the faces of the images using MTCNN to see if it would further increase recall, as we did with the gaze classifier. The resulting cropped dataset was significantly more balanced, with an approximately 80/20 split between no occlusion and occlusion images, although it contained fewer total images. After training the RGB model with the new dataset, accuracy remained similar at **99.70%**, but occlusion recall increased notably to **99.43%.**

For IR images, we also used a similar training strategy to the gaze classifier, but we did not crop the dataset. We achieved a test accuracy of **99.23%,** with occlusion recall score of **90.24%**.

## V. DRIVER IDENTIFICATION SYSTEM

The system uses RGB and IR cameras to capture the driver's face and employs machine-learning models for face detection and feature extraction. The RGB camera is the primary sensor, while the IR camera serves as a backup in low-light conditions or RGB camera malfunctions. This is represented in Fig. 5.

The captured image is processed by the MTCNN face detector. The largest detected face is sent to the feature extractor, which uses FaceNet [25] to create a unique embedding. If there was not face detected with RGB image, the system uses a CLAHE filter [26] on the IR image and repeats the process. If detection still fails, an "estimated BB" functionality which is using the previous face detection's bounding box extended by 20% to accommodate potential head movement. This bounding box is extracted from the same image type as the last detected face. If the "estimated BB" functionality is turned off, or if the system reaches a pre-defined maximum number of frames, the system will not return a detection and will wait for the next frame.

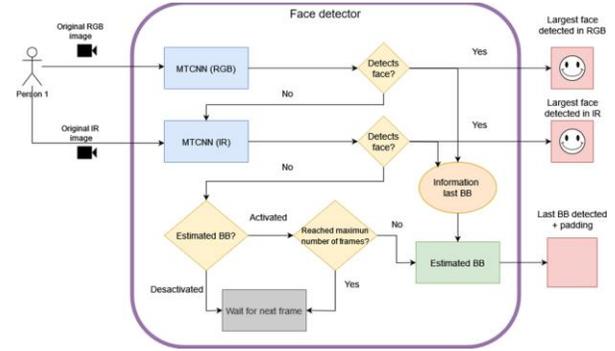

Fig. 5 Face detector module diagram

For re-identification, the system compares the new embedding with registered IDs in the database, matching if the cosine similarity exceeds 0.65 for RGB and 0.575 for IR images. These values were defined after experimenting and optimizing for better metrics. The system can also register new IDs automatically. The architecture of the system is illustrated in Fig. 6.

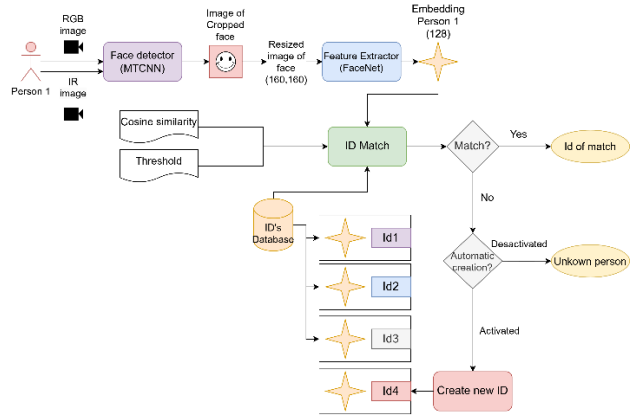

Fig. 6 Driver identification system diagram

Each identity in the database includes an ID number, name, and two embeddings (RGB and IR), allowing for identification in both lighting conditions. To enhance robustness, the embedding of every ID should represent more than one image of the person to handle different angles, illumination, and facial expressions. The final embedding is the mean of all images captured for one ID. During manual registration, three images with different head poses are captured to ensure varied embeddings. This method ensures that the embedding captures the driver's face from different angles, facilitating easier identification. To add a new embedding to an existing ID, we calculate the mean value of the new embedding and the existing one, considering the number of images it represents. For example, if the saved embedding represents four images, the new value is calculated as in (1). The new feature vector will then represent five images.

$$4/5 * oldEmbedding + 1/5 * newEmbedding \qquad (1)$$

To evaluate the Driver Identification System, we conducted experiments using the DMD dataset. The metrics used in these tests were accuracy, False Accept Rate (FAR), and False Reject Rate (FRR). FAR measures the rate at which unauthorized users are incorrectly accepted by the system, while FRR measures the rate at which authorized users are incorrectly rejected.

The tests utilized RGB images from the dataset, with ten individuals not included in the database and fifteen individuals included. One challenge encountered by the face detector module was the occurrence of false positives. To address this, we increased the confidence threshold of the MTCNN model, which evaluates potential face regions in the images. By setting the confidence value to 0.97, we effectively eliminated erroneous face detections without significantly reducing the number of valid detections. With the face detection functioning correctly, our approach of using mean embeddings to represent an ID achieved an accuracy of **99.38%,** a FAR of **0.81%**, and an FRR of **0.49%**.

## VI. PIPELINE AND IMPLEMENTATION

We created a pipeline that include all the algorithms described in this paper to build a DMS. The system uses a dual RGB and IR camera positioned behind the steering wheel to capture images of the driver. The three main components: driver identification, gaze estimation and occlusion detection are integrated and their data flow is described in the diagram of Fig. 7.

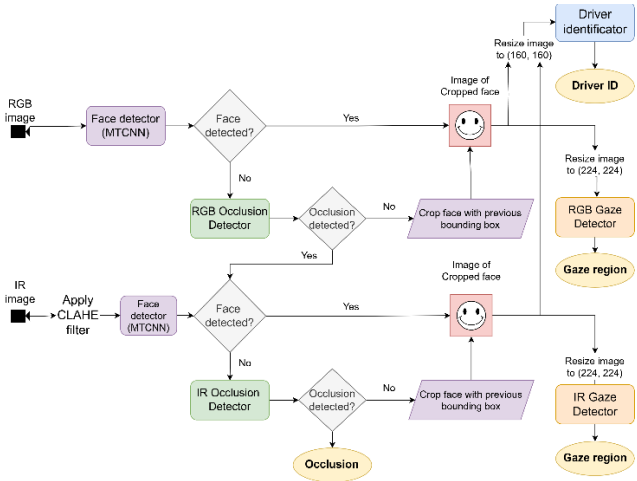

Fig. 7 DMS data flow diagram

We developed a pipeline integrating algorithms for a Driver Monitoring System (DMS) using dual RGB and IR cameras behind the steering wheel. The system includes driver identification, gaze estimation, and occlusion detection, as shown in Fig. 7. It starts with the RGB image passed to the face detector. If a face is detected, it is cropped and sent to the gaze classifier. If no face is detected, the image goes to the occlusion detector. If no occlusion is found, an "estimated bounding box" from the previous detection is used.

If occlusion is detected, the system switches to the IR image pipeline. Persistent occlusion with IR images indicates a failure or obstruction. The system alerts the driver and stops gaze estimation and identification.

The system continuously receives a stream of RGB and IR images. For each incoming image, the RGB image is processed by the algorithm; in parallel, the IR image is kept on hold for potential later use or for the driver identification system, which works with both image types. Driver identification runs periodically, assuming infrequent driver changes.

This pipeline primarily operates with RGB images, switching to IR images in low-light conditions or when the RGB sensor malfunctions. This switch becomes permanent if there is a consistent failure in face detection with the RGB image and no occlusion alert is emitted. The system reverts to the RGB flow after confirming consistent positive detections with RGB images. This design ensures a quick mode switch for tunnels and night conditions.

The models' quantitative results on the datasets provide a good representation of their general performance. However, we also tested these models in real-life scenarios to evaluate their performance in such environments. To test the gaze classifiers, subjects were instructed to look sequentially at each of the gaze regions. To test the occlusion classifiers, subjects were asked to occlude their faces in different ways to assess the models' robustness to various types of occlusions. While performance varied slightly between subjects, the RGB gaze classifier correctly classified all regions except for front right (region 5) which it almost always predicted as center mirror (region 4). The IR gaze classifier did not perform as well as its RGB counterpart, typically predicting left mirror (region 1), front (region 3), steering wheel (region 9), as shown in Fig. 9., and infotainment (region 8) correctly, but incorrectly predicting all other gaze zones. A graphical representation of the predictions, shown in Fig.8., was developed.

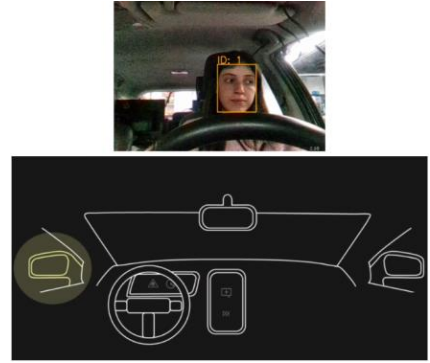

Fig. 8 Gaze classifier prediction of a subject looking at left mirror

The RGB occlusion classifier made accurate predictions in all scenarios we tested, including occluding the face with a hand or notebook, scratching eyes, scratching the forehead, and more. The IR classifier performed well in some cases, but was much less robust to different forms of occlusion. It often predicted occlusion when a hand was near the face but not occluding it, and did not predict occlusion when the subject was looking at the back seats, as shown in Fig. 9.

Overall, the RGB models performed quite well, demonstrating reliability and robustness, while the IR models performed weaker and were not as reliable or robust. These real-life results aligned closely with our expectations based on the models' performance on the test data.

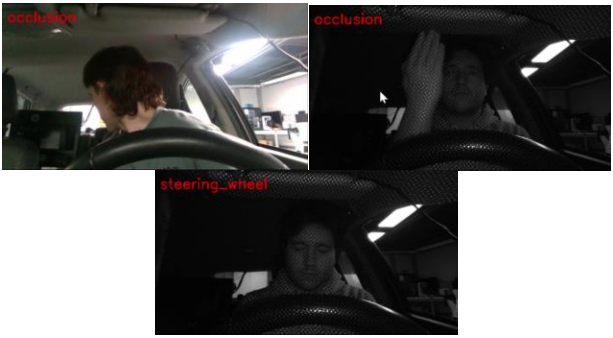

Fig. 9 Prediction examples of occlusion detection and gaze estimation

## VII. CONCLUSIONS

In this paper, we developed a driver gaze estimation by regions and driver identification pipeline that is aware of occlusions. We used the DMD dataset to train algorithms that could work with both RGB and IR images and integrate them into the pipeline. Our system is versatile, as it does not require calibration (for gaze estimation), and it can function in low-light conditions (e.g., nighttime, tunnels) and when one of the cameras is occluded or experiences technical issues. The system can also alert the driver when it cannot function properly, as recommended by EuroNCAP. The models were evaluated on datasets and in real-life scenarios, and the results showed that RGB models outperform IR models. This is likely due to the lack of pre-trained algorithms for IR images.

This research is a pioneering effort to introduce robust occlusion detection functionality into a DMS, which enhances its reliability. Future work could incorporate additional DMS functionalities, such as driver distraction detection. Further investigation into occlusion detection could also identify more specific causes of occlusions, as our algorithm was limited by the occlusions present in the DMD dataset. Expanding the dataset with more occlusion cases and IR data would be a valuable next step.